\PassOptionsToPackage{utf8}{inputenc}
\documentclass[nocrop]{bioinfo_arxiv}
\copyrightyear{2021} \pubyear{2021}

\access{Advance Access Publication Date: Day Month Year}
\appnotes{Original Paper}

\usepackage{multirow}
\usepackage{subcaption}
\usepackage{xspace}
\usepackage[table,xcdraw]{xcolor}

\usepackage{siunitx}
\newcommand{\predSource}[1]{\setlength{\fboxsep}{0.8pt}\setlength{\fboxrule}{0.9pt} \fcolorbox{white}{gray!30}{#1}}

\newcommand{\fscore}[1][1]{$F_{#1}$\xspace}
\newcommand{\ourmodel}{\mbox{\textit{CLIN-X}}\xspace}
\newcommand{\ourmodelES}{\mbox{\textit{CLIN-X}$_{ES}$}\xspace}
\newcommand{\ourmodelEN}{\mbox{\textit{CLIN-X}$_{EN}$}\xspace}

\newcommand{\corpusEnSix}{i2b2 2006\xspace}
\newcommand{\corpusEnTen}{i2b2 2010\xspace}
\newcommand{\corpusEnTwelve}{i2b2 2012\xspace}
\newcommand{\corpusEnTwelveCE}{i2b2 2012-C\xspace}
\newcommand{\corpusEnTwelveTime}{i2b2 2012-T\xspace}
\newcommand{\corpusEnFourteen}{i2b2 2014\xspace}

\newcommand{\corpusEsCantemist}{Cantemist\xspace}
\newcommand{\corpusEsMeddocan}{Meddocan\xspace}
\newcommand{\corpusEsMeddoprof}{Meddoprof\xspace}
\newcommand{\corpusEsMeddoprofNER}{Meddoprof-N\xspace}
\newcommand{\corpusEsMeddoprofClass}{Meddoprof-C\xspace}
\newcommand{\corpusEsPharmaconer}{Pharmaconer\xspace}

\begin{document}
\firstpage{1}

\subtitle{Data and Text Mining}

\title[\ourmodel]{\ourmodel: pre-trained language models and a study on cross-task transfer for concept extraction in the clinical domain}
\author[Lange \textit{et~al}.]{
Lukas Lange\,$^{\text{\sfb 1,2,}*}$, 
Heike Adel\,$^{\text{\sfb 1}}$, 
Jannik Strötgen\,$^{\text{\sfb 1}}$
and Dietrich Klakow\,$^{\text{\sfb 2}}$}
\address{
$^{\text{\sf 1}}$Bosch Center for Artificial Intelligence, Robert Bosch GmbH, Renningen, 71272, Germany and \\
$^{\text{\sf 2}}$Spoken Language Systems, Saarland University, Saarbrücken, 66111, Germany.
}

\corresp{$^\ast$To whom correspondence should be addressed.}

\abstract{\textbf{Motivation:} The field of natural language processing (NLP) has recently seen a large change towards using pre-trained language models for solving almost any task. Despite showing great improvements in benchmark datasets for various tasks, these models often perform sub-optimal in non-standard domains like the clinical domain where a large gap between pre-training documents and target documents is observed. In this paper, we aim at closing this gap with domain-specific training of the language model and we investigate its effect on a diverse set of downstream tasks and settings. \\
	\textbf{Results:} We introduce the pre-trained \ourmodel (Clinical XLM-R) language models and show how \ourmodel outperforms other pre-trained transformer models by a large margin for ten clinical concept extraction tasks from two languages. In addition, we demonstrate how the transformer model can be further improved with our proposed task- and language-agnostic model architecture based on ensembles over random splits and cross-sentence context. 
	Our studies in low-resource and transfer settings reveal stable model performance despite a lack of annotated data with improvements of up to 47 \fscore points when only 250 labeled sentences are available. Our results highlight the importance of specialized language models as \ourmodel for concept extraction in non-standard domains, 
	but also show that our task-agnostic model architecture is robust across the tested tasks and languages so that domain- or task-specific adaptations are not required. \\
	\textbf{Availability:} The \ourmodel language models and source code for fine-tuning and transferring the model are publicly available at \href{https://github.com/boschresearch/clin\_x}{https://github.com/boschresearch/clin\_x/} and the huggingface model hub. \\
	\textbf{Contact:} \href{Lukas.Lange@de.bosch.com}{Lukas.Lange@de.bosch.com}\\
}

\maketitle

\section{Introduction}

Collecting and understanding key clinical information, such as disorders, symptoms, drugs, etc., from electronic health records (EHRs) has wide-ranging applications within clinical practice and research~\citep{Leaman:Khare:JBI:2015,Wang:Wang:JBI:2018:Review}.
A better understanding of this information can, on the one hand, facilitate novel clinical studies, and, on the other hand, help practitioners to optimize clinical workflows.
However, free text is ubiquitous in EHRs. This leads to great difficulties in harvesting knowledge from EHRs.
Therefore, natural language processing (NLP) systems, especially information extraction components, play a critical role in extracting and encoding information of interest from clinical narratives, as this information can then be fed into downstream applications.
For example, the extraction of structured information from clinical narratives can help in decision making or drug repurposing \citep{Marimon:Gonzalez:MEDDOCAN:2019:Overview}.

However, information extraction in non-standard domains like the clinical domain is a challenging problem due to the large number of complex terms and unusual document structures \citep{Lee:Yoon:OxBio:2020:Biobert}. In addition, pre-trained language models (PLM) such as BERT~\citep{Devlin:Chang:NAACL:2018:Bert} that demonstrated superior performance for many NLP tasks are typically trained on standard domains, such as web texts, news articles or Wikipedia. Despite showing some robustness across languages and domains \citep{Conneau:Khandelwal:ACL:2020:Unsupervised} these models still achieve their best performance when applied to targets similar to their pre-training corpora which can limit their applicability in many situations \citep{Gururangan:Marasovi:ACL:2020:Dont}.  
One way to overcome this domain-gap is the adaptation of existing language models to the new target domain or training a new domain-specific model from scratch \citep{Beltagy:Lo:EMNLP:2019:Scibert,Lee:Yoon:OxBio:2020:Biobert}. 
Several recent works have shown that this kind of adaptation boosts performance for downstream tasks in non-standard domains by, e.g., pre-training with masked language modeling (MLM) objectives on documents from the target domain \citep{Weber:Muenchmeyer:OxBio:2019:HUNER,Aaseem:Kushi:IJCNN:2021:Bioalbert}.

While all the previous methods help to build high-performing model architectures, often there is also a lack of annotated data in the clinical domain which is usually needed for all deep-learning-based models. 
On the one hand, this domain has high requirements regarding the removal or masking of protected health information (PHI) of individuals \citep{Uzumer:Luo:i2b2:2006:Overview,Stubbs:Kotfile:i2b2:2014:Overview} which is particularly worthy of protection and can prevent data publication. On the other hand, information extraction tasks are often specific to their target domain and clinical concepts are only found very infrequently outside EHRs which limits reusability of existing resources. 
Possible solutions for the low-resource problem can be multi-task learning \citep{Khan:Ziyadi:2020:MtBioNER,Mulyar:Uzuner:2021:JAIMA:MtClinicalBert} or transfer Learning \citep{Lee:Dernoncourt:LREC:2018:Transfer,Peng:Yan:BioNLP:2019:TransferBio} across similar corpora from the clinical domain. 
However, transferring knowledge is particularly challenging in the clinical domain as biomedical NLP models have problems generalizing to new entities \citep{Kim:Kang:2021:BioGeneralize}.
Therefore, one has to carefully select the transfer sources \citep{Lange:Stroetgen:EMNLP:2021:Share}.

Over the last years, we have participated in a series of shared tasks on information extraction in the Spanish clinical domain \citep{Marimon:Gonzalez:MEDDOCAN:2019:Overview,Miranda:Farre:CANTEMIST:2020:Overview,Lima:Farre:MEDDOPROF:2021:Overview}. With our systems, we were able to outperform the other participants and won the competitions twice. The winning systems were task-agnostic and utilized domain-adapted language models and word embeddings \citep{Lange:Adel:NLNDE:2019:Meddocan}, as well as improved training routines for transformer models \citep{Lange:Adel:NLNDE:2021:Meddoprof}. 
Based on our findings and lessons learned during the competitions, we propose in this paper a robust model architecture and training procedure for concept extraction in the clinical domain that is task- and language-agnostic. 
We introduce a new Spanish clinical language model \ourmodelES (Clinical XLM-R) that outperforms existing transformer models on Spanish corpora and exemplifies the benefits of cross-language domain adaptation for English tasks as well.
For this, we perform a broad evaluation of ten clinical information extraction tasks from two languages (English and Spanish), including low-resource settings. 
Finally, we perform cross-task transfer experiments and show that this can boost performance by more than 47 \fscore points for few-shot training.
Our results demonstrate great and consistent improvements compared to standard transformer models across all tasks in both languages. 
We release both, \ourmodelES as well as its English counterpart \ourmodelEN.

\section{Approach}
\begin{figure}[!tpb]
	\centerline{\includegraphics[width=.45\textwidth]{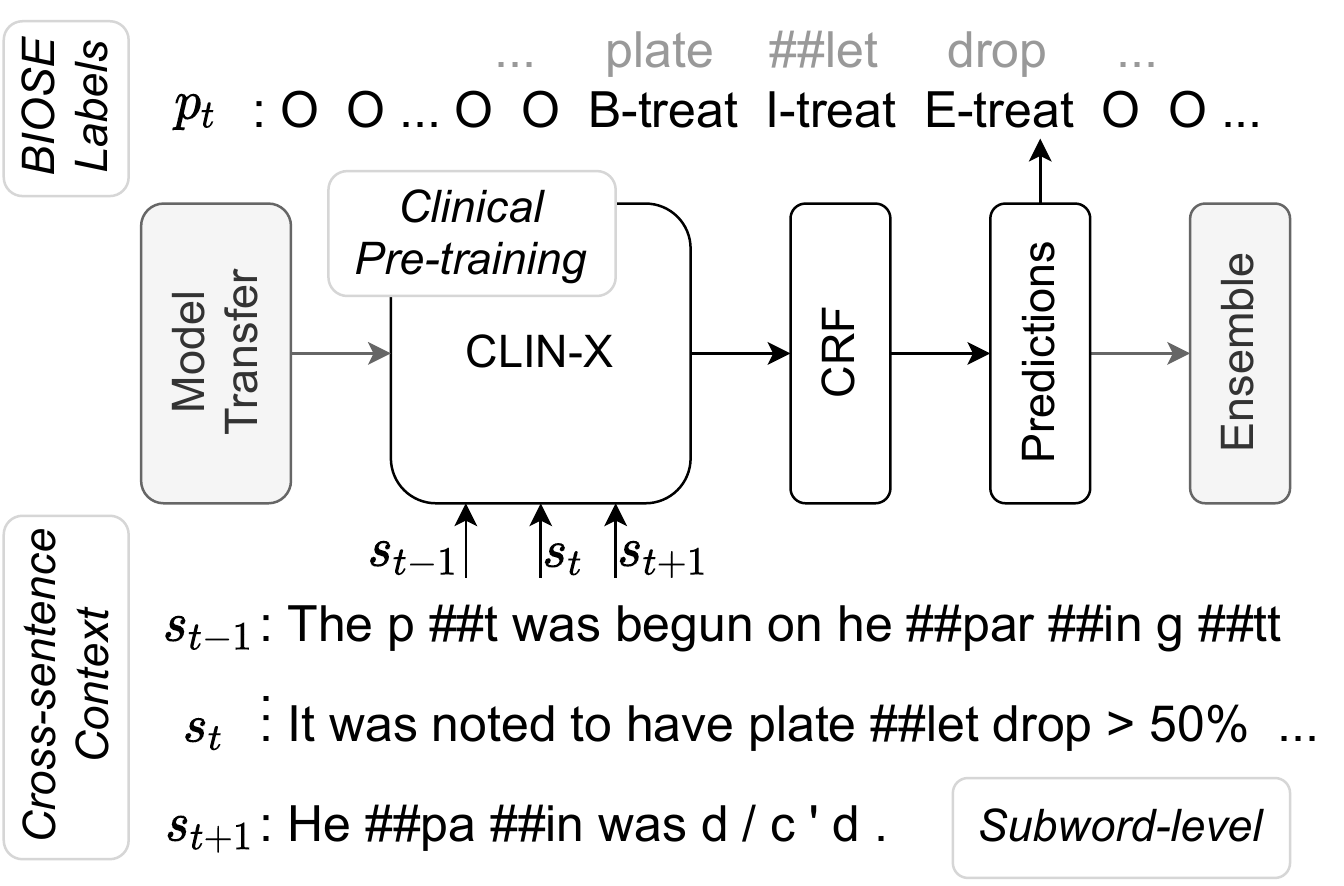}}
	\caption{Overview of the concept extraction pipeline based on \ourmodel and our model components for subword-based extraction with cross-sentence context, BIOSE labels, CRFs and model transfer. }\label{fig:model}
\end{figure}

In this paper, we introduce new pre-trained language models and propose a robust model architecture to perform concept extraction in the clinical domain for English and Spanish. The overall model architecture is shown in Figure~\ref{fig:model} and our proposed model components are highlighted. First, the input is computed on subword-level instead of the usual word-level, which eliminates the need for external tokenization. In addition, the input is enriched with its cross-sentence context to capture a wider document context. Second, the input is processed by a transformer model that is adapted to our target domain. Third, the model output is computed using a conditional random field (CRF) output layer to address long annotations.
Then, an ensemble over models trained on different training splits is computed that reduces variance and captures the complementary knowledge from all models. 
Finally, we experiment with cross-task model transfer to further improve the model in few-shot settings.

In summary, the contributions of this paper are as follows: 
\begin{itemize}
	
	\item We study the impact of domain-adaptive pre-training for clinical concept extraction for different embedding types and publish new language models that are adapted to the clinical domain. We show that this PLM outperforms other publicly available embeddings and models in our settings and we also show that cross-language domain adaptations works for English tasks as well.
	
	\item We perform a broad evaluation of ten clinical sequence labeling tasks across two languages, including low-resource and transfer settings. By this, we demonstrate how our methods can further boost already high-performing transformer models by using advanced training methods and effective changes in the architecture. 
	
	\item Our models outperform the state-of-the-art methods for clinical and biomedical concept extraction, as well as various other transformer models for all ten tasks. 
	
	\item We make our new domain-adapted \ourmodel language models and the source code for fine-tuning the concept extraction models using our methods publicly available.
	
\end{itemize}

\begin{methods}
	\section{Materials and Methods}
	In this section, we start with a brief description of the input representations. Then, we discuss our proposed architectural choices as well as the advanced training methods.

	\subsection{Input Representations for the Clinical Domain}
	State-of-the-art methods for concept extraction typically rely on word embeddings or language models as input representations. The standard approach is the pre-training of these models on large-scale unannotated datasets once and their reuse as powerful representations for many downstream applications \citep{Collobert:Weston:JMLR:2011:NLPfromScratch}.
	\cite{Phan:Sun:ACL:2019:ContextRepr} have shown that contextual information helps in particular in the medical domain, e.g., due to the high number of synonyms. Thus, we focus on the usage of contextualized embeddings in this work, which are most often retrieved from transformer language models nowadays. 
	This is either done with auto-regressive language modeling \citep{Peters:Neumann:NAACL:2018:Deep} 
	or masked language modeling \citep{Devlin:Chang:NAACL:2018:Bert}, which we use in this paper.

	\paragraph{Domain-specific embeddings.}
	A popular way to approach the challenges of NLP in non-standard domains is the inclusion of domain knowledge via domain-specific embeddings  \citep{Friedrich:Adel:ACL:2020:SOFC}. 
	For this, word embeddings or language models are pre-trained or further specialized on documents of the target domain. These embeddings can be used in downstream applications. 
	This kind of domain adaptation has shown great benefits in practice \citep{Gururangan:Marasovi:ACL:2020:Dont}, thus, we explore domain- and language-adaptive pre-training of transformer models in this paper.

	\paragraph{The \ourmodel pre-trained language model.}
	At the time of writing, there is no Spanish clinical transformer publicly available. 
	Thus, we train and publish the \ourmodelES language model. 
	The model is based on the multilingual XLM-R transformer, which was trained on 100 languages and showed superior performance in many different tasks across languages and can even outperform monolingual models in certain settings  \citep{Conneau:Khandelwal:ACL:2020:Unsupervised}.
	Even though XLM-R was pre-trained on 53GB of Spanish documents, this was only 2\% of the overall training data. To steer this model towards the Spanish clinical domain, we sample documents from the Scielo archive and the MeSpEn resources \citep{Villegas:Intxaurrondo:BIO:2018:Mespen}. The resulting corpus has a size of 790MB and is highly specific for our target setting. 
	We initialize \ourmodel using the pre-trained XLM-R weights and train masked language modeling (MLM) on the clinical corpus for 3 epochs which roughly corresponds to 32k steps.
	Nonetheless, this model is still multilingual and we demonstrate the positive impact of cross-language domain adaptation by applying this model to English tasks.
	\footnote{In addition to the Spanish \ourmodelES model, we release an English version \ourmodelEN trained on clinical Pubmed abstracts
		(850MB) filtered following \cite{Haynes:McKibbon:BMJ:2005:optimal} for a direct comparison of our methods in a monolingual setting. 
		This allows researchers and practitioners to address the English clinical domain with an out-of-the-box tailored model so that our transfer methods do not have to be applied.
		Pubmed is used with the courtesy of the U.S. National Library of Medicine.
	}

	\subsection{Concept Extraction Model}
	In the following, we describe the architectural choices we made compared to the standard transformer model for sequence labeling as proposed by \cite{Devlin:Chang:NAACL:2018:Bert}.

	\paragraph{Subword-level inputs.}
	Information extraction tasks are typically performed on the token level, while most transformers work on finer subwords instead. Thus, the input representations from transformers for tokens are either retrieved from the first subword or the average  \citep{Devlin:Chang:NAACL:2018:Bert}. In contrasts, we perform concept extraction directly on the subword level. By doing this, there is no need for external tokenization besides the subword segmentation of the transformer. Note that the usage of domain-specific subwords is still often beneficial compared to the general domain segmentation \citep{Beltagy:Lo:EMNLP:2019:Scibert,Lee:Yoon:OxBio:2020:Biobert}.

	\paragraph{Cross-sentence context.}
	Transformers are suited to incorporate information from a larger context.  \cite{Luoma:Ppysalo:COLING:2020:ContextNER} showed that context information from neighboring sentences has positive effects for named entity recognition on the general domain. \cite{Finkel:Dingare:BioNLP:2004:ContextBio} also showed the positive impact of context for clinical concept extraction. 
	We follow these approaches and add context information to the input similar to \cite{Schweter:Akbik:2020:flert}. We incorporate the context of 100 subwords to the left and right and use the document boundaries to set the context limits as all corpora are clearly separated in documents.

	\paragraph{Conditional Random Field Output. }
	As \cite{Kim:Kang:2021:BioGeneralize} have shown, entity recognition models in the biomedical domain tend to memorize training instances and their labels. This can result in incorrect label encodings as the model fails to generalize. A conditional random field \cite[CRF,][]{Lafferty:MaCallumg:ICML:2001:CRF} can constrain these incorrect sequences as the Viterbi algorithm is used for decoding. In addition, the CRF has advantages when it comes to long entities covering multiple tokens \citep{Lima:Farre:MEDDOPROF:2021:Overview} that appear frequently in the clinical domain.

	\subsection{Training on Data Splits.}
	Having a robust model architecture is a good starting point for NLP in the clinical domain. However, even more important might be the actual training procedure of the model. Thus, we discuss standard and random splits, as well as ensemble models over these splits in the following.

	\begin{figure}[!tpb]
		\centerline{\includegraphics[width=.45\textwidth]{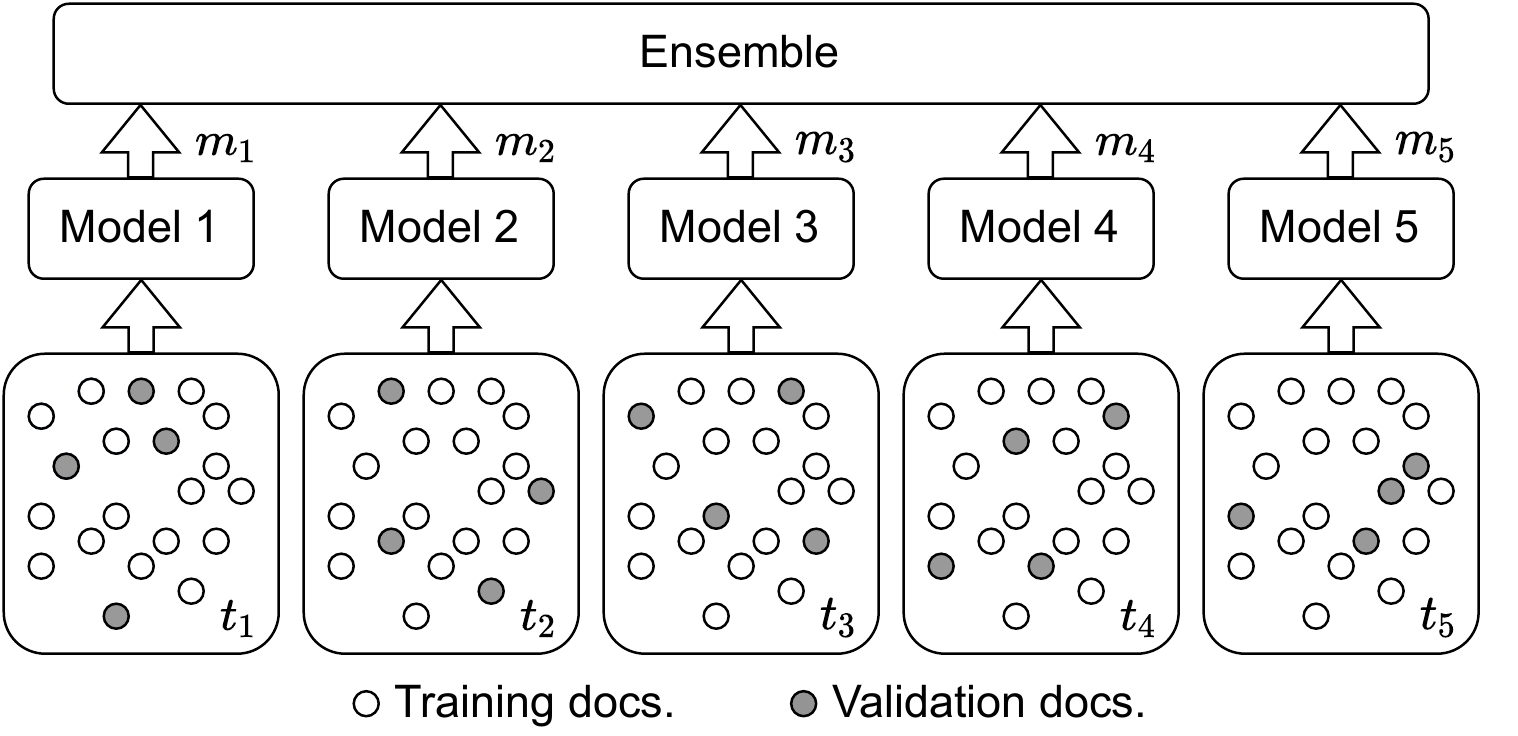}}
		\caption{Ensembles over different training splits splits. }\label{fig:splits}
	\end{figure}

	\paragraph{Standard splits.}
	Typically, each dataset is divided into training, development and test splits. The training split is used each epoch to train the model parameters and the best training epoch is selected based on the evaluation score on the development set. Finally, the held-out test set is used by the selected model to compute the final score. 
	These data splits are helpful to compare performances of different models on standardized data, however, using the standard training split without modifications may not result in optimal performance~\citep{Gorman:Bedrick:ACL:2019:Need}.

	\paragraph{Further random splits.}
	The training and development parts can be further randomly divided into $n$ separate parts. Then, $n-1$ parts can be used for training and one part as the validation set for early stopping similar to cross-fold validation. An ensemble based on models trained on the different data splits should be more powerful than the single models as each of them encodes complementary knowledge which helps to reduce variance and biases \citep{Clark:Yatskar:EMNLP:2019:Ensembles}. 
	In our experiments, we use $n=5$ so that we get 5 different settings with unique training sets and we train one model for each setting. 
	Note that we do not change or use the test set at all to ensure comparability to previous results.

	\paragraph{Training on all available instances.}
	Recent works sometimes finds that there is no need for a held-out development set and that these labeled instances might be better used during the training. For example, \cite{Luoma:Ppysalo:COLING:2020:ContextNER} have shown that training on the combined training and development sets boosts performance for named entity recognition remarkably. By this, the model has access to the most data during training and model selection is based on the training loss. 
	However, the training loss is not as meaningful as a stopping criterion and its hard to pick the best model checkpoint. We will compare to this method as an alternative to our split-based experiments.

	\subsection{Transfer Learning}
	Many NLP tasks suffer from a lack of labeled data. This includes non-standard domains like the clinical domain in particular. One solution to improve performance in these domains is the usage of resources from a related task in a transfer process. For example, \cite{Hofer:Kormilitzin:2018:FewShot} have shown that few-shot NER in the biomedical domain can be improved by transferring trained weights from a similar task. We perform a similar kind of model transfer by transferring the transformer to the new target. 
	
	However, not all transfer sources are actually useful as many can lead to negative transfer \citep{Lange:Stroetgen:EMNLP:2021:Share}. Thus, we first have to predict a suitable transfer source. We follow \cite{Lange:Stroetgen:EMNLP:2021:Share} and compute similarities between our datasets using their proposed model similarity measure. This has been shown to work well across different tasks and domains. The similarity between two models is computed based on the neural feature representations for the target datasets between two task-specific trained models. In our experiments, we study the effect of transfer from different sources in comparison to standard single-task training. Further, we will investigate this kind of transfer in low-resource settings, when the target task has only limited training resources.

	\paragraph{Ensembles over models. }
	In addition to the other methods, ensembling can be used to combine multiple model predictions into one. This ensemble is usually better than a single model -- in particular if the models or their training data differ to some degree. We either create ensembles by majority voting \citep{Clark:Yatskar:EMNLP:2019:Ensembles} of training runs that vary by their random seed (standard splits) or their training data (random splits).

	\section{Results}
	This section describes the experimental setup starting with tasks, datasets and implementation details, and discusses the results for our experiments.

	\begin{table}
		\footnotesize
		\centering
		\caption{Statistics of the Spanish datasets. }
		\begin{tabular}{lccc} \toprule
			\multirow{2}{*}{Corpus} & \multicolumn{3}{c}{Size (\#Sentences)} \\
			& Train & Dev & Test \\ \midrule 
			\corpusEsMeddocan \cite[]{Marimon:Gonzalez:MEDDOCAN:2019:Overview}       & 15,858 & 8,283 & 8,009 \\
			\corpusEsPharmaconer \cite[]{Gonzalez:Marimon:PHARMACONER:2019:Overview} & 8,582  & 4,016 & 4,184  \\
			\corpusEsCantemist \cite[]{Miranda:Farre:CANTEMIST:2020:Overview}        & 19,426 & 18,172 & 11,196 \\
			\corpusEsMeddoprof \cite[]{Lima:Farre:MEDDOPROF:2021:Overview}      & 51,350 & - & 10,008  \\
			\botrule
		\end{tabular}
		\label{tab:datasets-es}
	\end{table}
	
	\begin{table}
		\footnotesize
		\centering
		\caption{Statistics of the English datasets. }
		\begin{tabular}{lccc} \toprule
			\multirow{2}{*}{Corpus} & \multicolumn{3}{c}{Size (\#Sentences)} \\
			& Train & Dev & Test\\ \midrule 
			\corpusEnSix \cite[]{Uzumer:Luo:i2b2:2006:Overview}           & 51,429 & - & 18,770  \\
			\corpusEnTen \cite[]{Uzuner:South:i2b2:2010:Overview}         & 16,487 & - & 27,882  \\
			\corpusEnTwelve \cite[]{Sun:Rumshisky:i2b2:2012:Overview}    & 7,636  & - & 5,785  \\
			\corpusEnFourteen \cite[]{Stubbs:Kotfile:i2b2:2014:Overview}       & 52,026 & - & 33,317 \\ \botrule
		\end{tabular}
		\label{tab:datasets-en}
	\end{table}

	\subsection{Tasks and Datasets}\label{sub:tasks}
	Many datasets for natural language processing in specialized domains are published in the context of shared tasks -- competitions to evaluate different systems and approaches. 
	Besides English, the clinical domain is well addressed for Spanish, and there exists an active community of researchers for natural language processing of Spanish clinical texts. Thus, in the context of the IberLEF workshop series (Iberian Language Evaluation Forum), several shared tasks have been proposed by the Barcelona Supercomputing Center concerning concept extraction in the clinical domain
	\citep{Marimon:Gonzalez:MEDDOCAN:2019:Overview,Gonzalez:Marimon:PHARMACONER:2019:Overview,Miranda:Farre:CANTEMIST:2020:Overview,Lima:Farre:MEDDOPROF:2021:Overview}. 
	In addition to datasets of these shared tasks for Spanish, we consider four English datasets published during a series of shared tasks of the i2b2 project \citep{Uzumer:Luo:i2b2:2006:Overview,Uzuner:South:i2b2:2010:Overview,Sun:Rumshisky:i2b2:2012:Overview,Stubbs:Kotfile:i2b2:2014:Overview}. 
	Information on the dataset sizes are given in Table~\ref{tab:datasets-es} and \ref{tab:datasets-en} for Spanish and English, respectively. 
	Note that the Meddoprof and i2b2 2012 corpora consist of two different extraction tasks each. 
	Thus, we consider both tracks as separated tasks in this work resulting in a total of ten tasks. 
	Following the evaluations in the shared tasks, we use the strict micro \fscore for all datasets as evaluation metric.

	\begin{table}
		\centering
		\caption{Overview of different models averaged for the two languages (\fscore). Word embeddings are used in a RNN model similar to \cite{Akbik:Blythe:COLING:2018:Contextual}. Transformers are used with a classification layer similar to \cite{Devlin:Chang:NAACL:2018:Bert}. }
		\begin{tabular}{llcc} \toprule
			Pre-training Domain & Model & English & Spanish \\ \midrule
			\multirow{5}{*}{\begin{tabular}[l]{@{}l@{}}General \\ (e.g., Web, News, \\ Wikipedia, ...)\end{tabular}}
			& word2vec     & 80.26 & 78.20 \\
			& flair        & 85.15 & 80.28 \\
			& BERT (En)    & 85.34 & 77.78  \\ 
			& BETO (Es)    & 83.57 & 83.92  \\ 
			& XLM-R        & 87.13 & 83.87  \\ 
			\midrule
			\multirow{5}{*}{{Clinical}}
			& word2vec               & 80.98 & 79.72 \\
			& flair                  & 86.43 & 80.72 \\
			& ClinicalBERT (En)      & 85.76 & 76.94 \\
			& \ourmodelEN & \textbf{87.67} & 84.57  \\ 
			& \ourmodelES & 87.48 & \textbf{85.37}  \\ 
			\botrule
		\end{tabular}
		\label{tab:embeddings}
	\end{table}

	\subsection{Experimental Setup and Implementation Details}\label{sub:experiments}
	
	\paragraph{Masked Language Modeling.}
	We use eight NVIDIA V100 (32GB) GPUs for pre-training the \ourmodel models. The training takes less than 1 day with a batch size of 4 per device and a sequence length of up to 512 subwords. The models were trained with the huggingface trainer for MLM.

	\paragraph{Sequence Labeling.}
	The sequence labeling models were trained on single NVIDIA V100 GPUs up to 20 hours depending on the dataset size. The models were trained using the flair framework with the AdamW optimizer with an initial learning rate of \num{2.0e-5} and a batch size of 16 for 20 epochs. The model selection was performed on the development score if trained on standard or random splits or the training loss otherwise.

	\paragraph{Transfer and Low-Resource Experiments. }
	The median model according to the development score on the source dataset was taken for transfer and used for the initialization of the target model. Except for the initialization, the training was identical to the single task training. The low-resource settings were created by limiting the data splits to the first $n$ sentences without shuffling. The test set is not changed and remains identical.

	\subsection{Evaluation of Embeddings}\label{sub:eval-embeddings}
	The choice of input embeddings has a large impact on downstream performance and may be the most important factor. 
	Table~\ref{tab:embeddings} shows the average performance of several different embeddings and transformer models for the two languages. As expected, the monolingual transformers (BERT, BETO) excel at their target language, but cannot compete with multilingual models (mBERT, XLM-R) when applied to an unseen language. 
	The lower part of Table~\ref{tab:embeddings} lists domain-specific variants of the embeddings which are generally more powerful in our domain-specific setting. 
	We see that our \ourmodel models perform best for their respective languages. Furthermore, the \ourmodelES performs almost as well as the \ourmodelEN model on the English datasets, for which it was not explicitly trained. This shows, that the domain adaptation of multilingual models can also help for texts from other languages of the same domain. Due to \ourmodelES stable performance across all tasks and languages, we will use this model for the following ablations and transfer experiments.

	\begin{table}
		\centering
		\caption{Comparison of training splits with our model architecture and ablation study of the model components averaged for each language (\fscore). }
		\begin{tabular}{lllcc} \toprule
			& & Method      & English & Spanish \\ \midrule
			& & All           & 87.83 & 86.46 \\ \midrule
			\multirow{3}{*}{\rotatebox{90}{Standard}} & 
			\multirow{3}{*}{\rotatebox{90}{Splits}}
			& Median model  & 87.63 & 85.16 \\
			& & Best model    & 87.85 & 85.99 \\ 
			& & Ensemble      & 87.95 & 86.06 \\ \midrule
			\multirow{3}{*}{\rotatebox{90}{Random}} &
			\multirow{3}{*}{\rotatebox{90}{Splits}}
			& Median model  & 87.69 & 86.17 \\
			& & Best model    & 88.31 & 86.85 \\ 
			& & Ensemble      & \textbf{88.78} & \textbf{88.15} \\ \midrule
			\multirow{4}{*}{\rotatebox{90}{Ablation}} &
			\multirow{4}{*}{\rotatebox{90}{Study}}
			& -- BIOSE Labels  & 88.52   & 87.13 \\ 
			& & -- CRF           & 88.38   & 85.95 \\ 
			& & -- Context       & 87.83   & 86.84 \\
			& & -- Subword NER   & 87.38   & 86.81 \\ \botrule
		\end{tabular}
		\label{tab:splits}
	\end{table}
	
	\begin{table}
		\centering
		\caption{Cross-task transfer results for few-shot settings for the English corpora (\fscore). The \predSource{predicted transfer source} and the \textbf{best} models are highlighted.}
		\begin{tabular}{clcccccc} \toprule
			& & \multicolumn{6}{c}{\# training sentences} \\
			Tgt. & Src. / Setting & 250 & 500 & 1000 & 2500 & 7500 & All \\ 
			\midrule
			\multirow{5}{*}{\rotatebox{90}{\corpusEnSix}}
			& No Transfer & 71.24 & 81.06 & 84.15 & 95.49 & 96.89 & 98.34 \\
			& \corpusEnTen & 81.55 & 90.38 & 89.09 & 95.61 & 97.47 & 96.88 \\
			& \corpusEnTwelveCE & 79.28 & 86.5 & 88.71 & 96.75 & 97.92 & 98.23 \\
			& \corpusEnTwelveTime & 71.58 & 80.31 & 83.29 & 95.87 & \textbf{97.97} & 97.41 \\
			& \cellcolor[HTML]{D3D3D3}\corpusEnFourteen & \cellcolor[HTML]{D3D3D3}\textbf{87.52} & \cellcolor[HTML]{D3D3D3}\textbf{90.86} & \cellcolor[HTML]{D3D3D3}\textbf{91.87} & \cellcolor[HTML]{D3D3D3}\textbf{97.11} & \cellcolor[HTML]{D3D3D3}97.95 & \cellcolor[HTML]{D3D3D3}\textbf{98.50} \\ 
			\midrule
			\multirow{5}{*}{\rotatebox{90}{\corpusEnTen}}
			& No Transfer & 65.38 & 74.96 & 82.59 & 85.54 & 88.48 & 89.10 \\
			& \corpusEnSix & 68.90 & 78.32 & 82.07 & 85.70 & 87.95 & 88.69 \\
			& \cellcolor[HTML]{D3D3D3}\corpusEnTwelveCE & \cellcolor[HTML]{D3D3D3}\textbf{83.99} & \cellcolor[HTML]{D3D3D3}\textbf{86.25} & \cellcolor[HTML]{D3D3D3}\textbf{86.88} & \cellcolor[HTML]{D3D3D3}\textbf{88.46} & \cellcolor[HTML]{D3D3D3}\textbf{89.34} & \cellcolor[HTML]{D3D3D3}\textbf{89.74} \\
			& \corpusEnTwelveTime & 69.49 & 74.92 & 81.31 & 85.35 & 88.25 & 88.65 \\
			& \corpusEnFourteen & 72.05 & 79.11 & 82.49 & 85.54 & 87.69 & 88.80 \\
			\midrule
			\multirow{5}{*}{\rotatebox{90}{\corpusEnTwelveCE}}
			& No Transfer & 69.09 & 73.21 & 75.70 & 78.03 & 80.36 & 80.42 \\
			& \corpusEnSix & 68.83 & 72.14 & 75.34 & 77.86 & 79.25 & 80.15 \\
			& \cellcolor[HTML]{D3D3D3}\corpusEnTen & \cellcolor[HTML]{D3D3D3}\textbf{76.39} & \cellcolor[HTML]{D3D3D3}\textbf{77.98} & \cellcolor[HTML]{D3D3D3}\textbf{79.44} & \cellcolor[HTML]{D3D3D3}\textbf{80.90} & \cellcolor[HTML]{D3D3D3}\textbf{81.65} & \cellcolor[HTML]{D3D3D3}\textbf{80.93} \\
			& \corpusEnTwelveTime & 65.30 & 69.61 & 73.30 & 75.88 & 80.25 & 80.12 \\
			& \corpusEnFourteen & 68.67 & 72.56 & 75.39 & 77.96 & 79.98 & 79.83 \\
			\midrule
			\multirow{5}{*}{\rotatebox{90}{\corpusEnTwelveTime}}
			& No Transfer & 67.49 & 72.67 & 75.44 & 78.00 & 78.33 & 78.48 \\
			& \corpusEnSix & 68.57 & 72.49 & 74.34 & 77.73 & 78.43 & 78.34 \\
			& \cellcolor[HTML]{D3D3D3}\corpusEnTen & \cellcolor[HTML]{D3D3D3}68.10 & \cellcolor[HTML]{D3D3D3}74.04 & \cellcolor[HTML]{D3D3D3}\textbf{78.01} & \cellcolor[HTML]{D3D3D3}\textbf{78.98} & \cellcolor[HTML]{D3D3D3}\textbf{79.29} & \cellcolor[HTML]{D3D3D3}79.60 \\
			& \corpusEnTwelveCE & \textbf{70.17} & \textbf{75.04} & 76.36 & 78.12 & 78.54 & \textbf{80.03} \\
			& \corpusEnFourteen & 69.44 & 72.66 & 75.04 & 77.88 & 78.86 & 79.36 \\
			\midrule
			\multirow{5}{*}{\rotatebox{90}{\corpusEnFourteen}}
			& No Transfer & 64.96 & 81.61 & 85.74 & 92.70 & 96.08 & \textbf{97.62} \\
			& \cellcolor[HTML]{D3D3D3}\corpusEnSix & \cellcolor[HTML]{D3D3D3}\textbf{81.50} & \cellcolor[HTML]{D3D3D3}\textbf{85.76} & \cellcolor[HTML]{D3D3D3}\textbf{88.96} & \cellcolor[HTML]{D3D3D3}\textbf{93.51} & \cellcolor[HTML]{D3D3D3}96.04 & \cellcolor[HTML]{D3D3D3}97.46 \\
			& \corpusEnTen & 71.72 & 83.55 & 87.81 & 93.18 & \textbf{96.14} & 97.17 \\
			& \corpusEnTwelveCE & 71.24 & 82.97 & 87.09 & 93.15 & 96.13 & 97.33 \\
			& \corpusEnTwelveTime & 69.12 & 81.25 & 85.08 & 91.35 & 96.02 & 97.00 \\ 
			\botrule
		\end{tabular}
		\label{tab:low-resource-en}
	\end{table}

	\begin{table}
		\centering
		\caption{Cross-task transfer results for few-shot settings for the Spanish corpora (\fscore). The \predSource{predicted transfer source} and the \textbf{best} models are highlighted.}
		\begin{tabular}{clcccccc} \toprule
			& & \multicolumn{6}{c}{\# training sentences} \\
			Tgt. & Src. / Setting & 250 & 500 & 1000 & 2500 & 7500 & All \\ 
			\midrule
			\multirow{5}{*}{\rotatebox{90}{\corpusEsCantemist}}
			& No Transfer & 51.68 & 59.00 & 67.35 & 77.15 & \textbf{84.10} & \textbf{88.24} \\
			& \cellcolor[HTML]{D3D3D3}\corpusEsMeddocan & \cellcolor[HTML]{D3D3D3}\textbf{56.48} & \cellcolor[HTML]{D3D3D3}59.51 & \cellcolor[HTML]{D3D3D3}\textbf{69.33} & \cellcolor[HTML]{D3D3D3}76.57 & \cellcolor[HTML]{D3D3D3}83.43 & \cellcolor[HTML]{D3D3D3}88.00 \\
			& \corpusEsMeddoprofNER & 52.06 & \textbf{59.26} & 67.18 & \textbf{77.27} & 83.05 & 87.74 \\
			& \corpusEsMeddoprofClass & 53.94 & 55.41 & 65.71 & 76.65 & 83.20 & 88.00 \\
			& \corpusEsPharmaconer & 55.53 & 59.14 & 66.78 & 76.44 & 83.39 & 87.95 \\
			\midrule
			\multirow{5}{*}{\rotatebox{90}{\corpusEsMeddocan}}
			& No Transfer & 84.00 & 92.01 & 95.28 & 96.48 & 97.20 & \textbf{98.00} \\
			& \corpusEsCantemist & 83.61 & 89.36 & 95.35 & 96.75 & 97.43 & 97.57 \\
			& \corpusEsMeddoprofNER & 86.99 & 92.77 & 93.55 & 96.15 & 97.01 & 97.66 \\
			& \corpusEsMeddoprofClass & 88.70 & 93.76 & 95.03 & 96.32 & 97.35 & 97.73 \\
			& \cellcolor[HTML]{D3D3D3}\corpusEsPharmaconer & \cellcolor[HTML]{D3D3D3}\textbf{92.74} & \cellcolor[HTML]{D3D3D3}\textbf{94.30} & \cellcolor[HTML]{D3D3D3}\textbf{96.16} & \cellcolor[HTML]{D3D3D3}\textbf{96.84} & \cellcolor[HTML]{D3D3D3}\textbf{97.49} & \cellcolor[HTML]{D3D3D3}97.65 \\
			\midrule
			\multirow{5}{*}{\rotatebox{90}{\corpusEsMeddoprofNER}}
			& No Transfer & 13.99 & 44.28 & 51.24 & 58.95 & 72.54 & 81.68 \\
			& \corpusEsCantemist & 10.01 & 38.41 & 50.64 & 62.66 & 71.74 & 79.77 \\
			& \corpusEsMeddocan & 16.39 & 45.30 & 52.89 & 62.25 & 73.30 & 81.38 \\
			& \cellcolor[HTML]{D3D3D3}\corpusEsMeddoprofClass & \cellcolor[HTML]{D3D3D3}\textbf{61.29} & \cellcolor[HTML]{D3D3D3}\textbf{68.37} & \cellcolor[HTML]{D3D3D3}\textbf{72.83} & \cellcolor[HTML]{D3D3D3}\textbf{72.88} & \cellcolor[HTML]{D3D3D3}\textbf{78.04} & \cellcolor[HTML]{D3D3D3}\textbf{81.88} \\
			& \corpusEsPharmaconer & 23.72 & 44.91 & 52.90 & 60.53 & 73.35 & 81.07 \\
			\midrule
			\multirow{5}{*}{\rotatebox{90}{\corpusEsMeddoprofClass}}
			& No Transfer & 16.46 & 24.28 & 47.67 & 54.66 & 68.68 & \textbf{80.54} \\
			& \corpusEsCantemist & 10.99 & 29.73 & 49.20 & 52.75 & 66.57 & 78.76 \\
			& \corpusEsMeddocan & 31.83 & 38.01 & 53.80 & 56.46 & 69.98 & 79.33 \\
			& \cellcolor[HTML]{D3D3D3}\corpusEsMeddoprofNER & \cellcolor[HTML]{D3D3D3}\textbf{57.46} & \cellcolor[HTML]{D3D3D3}\textbf{57.70} & \cellcolor[HTML]{D3D3D3}\textbf{61.56} & \cellcolor[HTML]{D3D3D3}\textbf{64.92} & \cellcolor[HTML]{D3D3D3}\textbf{72.37} & \cellcolor[HTML]{D3D3D3}79.38 \\
			& \corpusEsPharmaconer & 22.61 & 35.15 & 50.50 & 53.49 & 69.59 & 79.08 \\
			\midrule
			\multirow{5}{*}{\rotatebox{90}{\corpusEsPharmaconer}}
			& Sinlge-Task & 67.71 & 76.38 & 81.32 & 87.68 & 91.31 & 92.27 \\
			& Cantemist & 60.34 & 71.77 & 79.45 & 86.77 & 90.61 & \textbf{92.35} \\
			& \cellcolor[HTML]{D3D3D3}Meddocan & \cellcolor[HTML]{D3D3D3}\textbf{74.48} & \cellcolor[HTML]{D3D3D3}76.02 & \cellcolor[HTML]{D3D3D3}\textbf{82.79} & \cellcolor[HTML]{D3D3D3}\textbf{88.39} & \cellcolor[HTML]{D3D3D3}\textbf{91.49} & \cellcolor[HTML]{D3D3D3}92.27 \\
			& \corpusEsMeddoprofNER & 69.48 & \textbf{76.44} & 78.73 & 88.60 & 92.02 & 91.98 \\
			& \corpusEsMeddoprofClass & 69.25 & 74.15 & 80.13 & 88.27 & 91.80 & 92.29 \\
			\botrule
		\end{tabular}
		\label{tab:low-resource-es}
	\end{table}

	\newcommand{\cbr}{$^{\dagger}$}
	\begin{table*}
		\centering
		\footnotesize
		\caption{Comparison to baseline systems and state-of-the-art results  (\fscore). 
			We highlight statistically significant differences between \ourmodelES +OurArchitecture with and without transfer following the significant codes of R: 
			*** $p$-value $\leq 0.001$; ** $p$-value $< 0.01$; * $p$-value $< 0.05$;
			\cbr highlights our ClinicalBERT results. }
		\begin{tabular}{cl|ccccc|ccccc} \toprule
			& & \multicolumn{5}{c|}{English (i2b2)} & \multicolumn{5}{c}{Spanish} \\
			& Model & 2006 & 2010 & 2012-C & 2012-T & 2014 &
			\corpusEsCantemist & \corpusEsMeddocan & M.prof-N & M.prof-C & Pharma. \\ \midrule 
			\multirow{3}{*}{\rotatebox{90}{}} 
			& BERT/BETO (monolingual) & 94.80 & 85.25 & 76.51 & 75.28 & 94.86 & 81.30 & 96.81 & 79.19 & 74.59 & 87.70 \\
			& BERT (multilingual)     & 94.79 & 84.91 & 76.01 & 76.56 & 95.34 & 80.94 & 96.30 & 76.39 & 71.84 & 86.98 \\
			& XLM-R (multilingual)    & 96.72 & 87.54 & 79.63 & 75.36 & 96.39 & 82.17 & 96.76 & 77.44 & 74.05 & 88.92 \\
			& HunFlair (monolingual)  & 93.48 & 86.70 & 78.52 & 77.16 & 95.90 & 83.80 & 96.50 & 75.16 & 70.01 & 88.40 \\ \midrule
			& ClinicalBERT & 94.8   & 87.8   & 78.9   & 76.58\cbr     & 93.0  & 77.18\cbr  & 94.63\cbr & 65.74\cbr & 62.85\cbr & 84.32\cbr \\
			& NLNDE         & -      & -      & -      & -     & -     & 85.3  & 96.96 & 81.8  & 79.3  & 88.6  \\ \midrule
			& \ourmodelEN & 96.25 & 88.10 & 79.58 & 77.70 & 96.73 & 82.80 & 97.08 & 78.62 & 75.05 & 89.33 \\
			& \ourmodelES & 95.49 & 87.94 & 79.58 & 77.57 & 96.80 & 83.22 & 97.08 & 79.54 & 76.95 & 90.05 \\ \midrule
			& \ourmodelEN +OurArchitecture & 98.49 & 89.23 & 80.62 & 78.50 & 97.60 & 87.72 & 97.57 & 81.36 & 78.53 & \textbf{92.36} \\
			& \ourmodelES +OurArchitecture & 98.30 & 89.10 & 80.42 & 78.48 & \textbf{97.62}* & \textbf{88.24} & \textbf{98.00} & 81.68 & \textbf{80.54} & 92.27 \\ \midrule
			& \ourmodelES +OurArchitecture +Transfer  & \textbf{98.50}* & \textbf{89.74}*** & \textbf{80.93}** & \textbf{79.60}* & 97.46 & 88.00 & 97.65 & \textbf{81.88} & 79.38 & 92.27 \\ \botrule
		\end{tabular}
		\label{tab:final}
	\end{table*}

	\subsection{Evaluation of Training Methods}\label{sub:eval-train}
	The foundation for all following concept extraction models is the \ourmodelES transformer, as it has shown robust results across all tasks. 
	For comparison to fixed standard splits, we train the models on different random splits. 
	We see in Table~\ref{tab:splits} that in particular ensembles over random splits are a lot better than the standard splits and also all training instances. 
	While the median performance is roughly similar for all methods, the random splits offer a lot more variety in training instances and allow for better maximum performance models. Thus, the ensemble based on random splits achieves also much higher numbers.

	\subsection{Evaluation of Concept Extraction Models}\label{sub:eval-model}
	The lower part of Table~\ref{tab:splits} lists an ablation study of our individual model components. 
	For example, adding cross-sentence context to the transformers boosts performance across all tasks by 0.5 F1 on average. 
	Performing concept extraction on the subword level helps even further. This is particularly helpful considering that no external tokenization is needed, which can be challenging in the clinical domain \citep{Lange:Adel:NLNDE:2020:Cantemist}.
	The CRF helps for both languages, though the differences are larger for Spanish, as the two MEDDOPROF tasks have particularly long annotations (2.53 tokens per annotation on average). 
	The same holds for the BIOSE labels, that have the smallest impact of all components, but consistently improve upon the standard BIO labels. 
	As each of our proposed methods improves the transformer even further, we use the combination of all methods in the following as our model architecture.

	\subsection{Evaluation of Transfer Learning}\label{sub:eval-transfer}
	In addition to the training based on random splits, we explore the effects of transfer learning. For this, we simulate low-resource settings where we limit the annotated data of the target dataset between 250 labeled sentences up to 7500 sentences, roughly the size of the smallest corpus. The results are given in Table~\ref{tab:low-resource-en} and Table~\ref{tab:low-resource-es} for English and Spanish, respectively. 
	
	Large positive transfer happens in most settings, particularly for the low-resource settings with up to (+47.3 \fscore points) for \corpusEsMeddoprof when only 250 labeled sentences are available. The improvements in the full data scenario are below 1 F1. 
	However, there is also negative transfer, in particular using \corpusEnTwelveTime and \corpusEsCantemist datasets as transfer sources often result in negative transfer. The source selection is also crucial in low-resource scenarios, as not every source is equally beneficial. 
	Using the model similarity measure from \cite{Lange:Stroetgen:EMNLP:2021:Share} we are able to predict good transfer sources in all settings; often the best source is selected.

	\subsection{Comparison to State-of-the-Art Models}\label{sub:eval-sota}
	As our results demonstrate, we have proposed a robust model for the clinical domain that works well across the different tasks in both languages. 
	Finally, we compare \ourmodel to various transformer models as introduced earlier. We also compare to HunFlair \citep{Weber:Saenger:OxBio:2021:HunFlair}, the current state-of-the-art for concept extraction in the biomedical domain. We use their model architecture based on clinical flair and fasttext embeddings and train models accordingly on our datasets. 
	In addition, we compare to our NLNDE submissions for the Spanish shared tasks and the ClinicalBERT by \cite{Alsentzer:Murphy:CNLP:2019:ClinicalBert} for the English datasets.
	
	The results for each task are shown in Table~\ref{tab:final}.
	The \ourmodel language models in combination with our model architecture outperform the other transformers and HunFlair by a large margin. \ourmodel is able to utilize the domain knowledge obtained from the additional pre-training with further improvements from the ensembling over random splits. Even though \ourmodel works best in combination with our model architecture, \ourmodel based on the standard transformer architecture with a single classification layer already outperforms the existing models on 8 out of 10 tasks.
	
	We tested statistical significance between \ourmodelES with and without transfer learning -- highlighted with asterisks in Table~\ref{tab:final}. 
	We find that all differences for English are significant, while only one difference for Spanish is significant. This might indicate the complementary relationship of domain adaptation and model transfer learning. As \ourmodel was explicitly adapted to Spanish, additional transfer is not necessary in high-resource settings. 
	In contrast, the cross-language domain adaptation for English can still be improved with transfer from related sources, where \ourmodelES+Transfer has also notably higher performances in 3 out of 5 settings compared to \ourmodelEN which is adapted to English.

\end{methods}

\section{Conclusion}\label{sec:conclusion}
In this paper, we described the newly pre-trained \ourmodel language  models for the clinical domain. We have shown that \ourmodel sets the new state the of the art results for ten clinical concept extraction tasks in two languages. We demonstrated the positive impact of other model components, such as ensembles over random splits and cross-sentence context and we have studied the effects of cross-task transfer learning from different clinical corpora. Using a model similarity measure, we found good transfer sources for almost all datasets in general and for low-resource scenarios in particular. 
We are convinced that the new \ourmodel language models will help boosting performance for various Spanish and English clinical information extraction tasks with our or other model architectures.

\bibliographystyle{natbib}
\bibliography{document}

\begin{thebibliography}{}

\bibitem[Akbik {\em et~al.}(2018)Akbik, Blythe, and
  Vollgraf]{Akbik:Blythe:COLING:2018:Contextual}
Akbik, A.  {\em et~al.} (2018).
\newblock Contextual string embeddings for sequence labeling.
\newblock In {\em Proceedings of the 27th International Conference on
  Computational Linguistics\/}, pages 1638--1649, Santa Fe, New Mexico, USA.
  ACL.

\bibitem[Alsentzer {\em et~al.}(2019)Alsentzer, Murphy, Boag, Weng, Jindi,
  Naumann, and McDermott]{Alsentzer:Murphy:CNLP:2019:ClinicalBert}
Alsentzer, E.  {\em et~al.} (2019).
\newblock Publicly available clinical {BERT} embeddings.
\newblock In {\em Proceedings of the 2nd Clinical Natural Language Processing
  Workshop (Clin-NLP)\/}, pages 72--78, Minneapolis, Minnesota, USA. ACL.

\bibitem[Beltagy {\em et~al.}(2019)Beltagy, Lo, and
  Cohan]{Beltagy:Lo:EMNLP:2019:Scibert}
Beltagy, I.  {\em et~al.} (2019).
\newblock {S}ci{BERT}: A pretrained language model for scientific text.
\newblock In {\em Proceedings of the 2019 Conference on Empirical Methods in
  Natural Language Processing and the 9th International Joint Conference on
  Natural Language Processing (EMNLP-IJCNLP)\/}, pages 3615--3620, Hong Kong,
  China. ACL.

\bibitem[Clark {\em et~al.}(2019)Clark, Yatskar, and
  Zettlemoyer]{Clark:Yatskar:EMNLP:2019:Ensembles}
Clark, C.  {\em et~al.} (2019).
\newblock Don{'}t take the easy way out: Ensemble based methods for avoiding
  known dataset biases.
\newblock In {\em Proceedings of the 2019 Conference on Empirical Methods in
  Natural Language Processing and the 9th International Joint Conference on
  Natural Language Processing (EMNLP-IJCNLP)\/}, pages 4069--4082, Hong Kong,
  China. ACL.

\bibitem[Collobert {\em et~al.}(2011)Collobert, Weston, Bottou, Karlen,
  Kavukcuoglu, and Kuksa]{Collobert:Weston:JMLR:2011:NLPfromScratch}
Collobert, R.  {\em et~al.} (2011).
\newblock Natural language processing (almost) from scratch.
\newblock {\em J. Mach. Learn. Res.}, {\bf 12}, 2493--2537.

\bibitem[Conneau {\em et~al.}(2020)Conneau, Khandelwal, Goyal, Chaudhary,
  Wenzek, Guzm{\'a}n, Grave, Ott, Zettlemoyer, and
  Stoyanov]{Conneau:Khandelwal:ACL:2020:Unsupervised}
Conneau, A.  {\em et~al.} (2020).
\newblock Unsupervised cross-lingual representation learning at scale.
\newblock In {\em Proceedings of the 58th Annual Meeting of the Association for
  Computational Linguistics (ACL)\/}, pages 8440--8451, Online. ACL.

\bibitem[Devlin {\em et~al.}(2019)Devlin, Chang, Lee, and
  Toutanova]{Devlin:Chang:NAACL:2018:Bert}
Devlin, J.  {\em et~al.} (2019).
\newblock {BERT}: Pre-training of deep bidirectional transformers for language
  understanding.
\newblock In {\em Proceedings of the 2019 Conference of the North {A}merican
  Chapter of the Association for Computational Linguistics: Human Language
  Technologies (NAACL-HLT)\/}, pages 4171--4186, Minneapolis, Minnesota. ACL.

\bibitem[Finkel {\em et~al.}(2004)Finkel, Dingare, Nguyen, Nissim, Manning, and
  Sinclair]{Finkel:Dingare:BioNLP:2004:ContextBio}
Finkel, J.  {\em et~al.} (2004).
\newblock Exploiting context for biomedical entity recognition: From syntax to
  the web.
\newblock In {\em Proceedings of the International Joint Workshop on Natural
  Language Processing in Biomedicine and its Applications (NLPBA/BioNLP)\/},
  pages 91--94, Geneva, Switzerland. International Committee on Computational
  Linguistics.

\bibitem[Friedrich {\em et~al.}(2020)Friedrich, Adel, Tomazic, Hingerl,
  Benteau, Marusczyk, and Lange]{Friedrich:Adel:ACL:2020:SOFC}
Friedrich, A.  {\em et~al.} (2020).
\newblock The {SOFC}-exp corpus and neural approaches to information extraction
  in the materials science domain.
\newblock In {\em Proceedings of the 58th Annual Meeting of the Association for
  Computational Linguistics (ACL)\/}, pages 1255--1268, Online. ACL.

\bibitem[Gonzalez-Agirre {\em et~al.}(2019)Gonzalez-Agirre, Marimon,
  Intxaurrondo, Rabal, Villegas, and
  Krallinger]{Gonzalez:Marimon:PHARMACONER:2019:Overview}
Gonzalez-Agirre, A.  {\em et~al.} (2019).
\newblock {P}harma{C}o{NER}: Pharmacological substances, compounds and proteins
  named entity recognition track.
\newblock In {\em Proceedings of The 5th Workshop on BioNLP Open Shared Tasks
  (BioNLP-OST)\/}, pages 1--10, Hong Kong, China. ACL.

\bibitem[Gorman and Bedrick(2019)Gorman and
  Bedrick]{Gorman:Bedrick:ACL:2019:Need}
Gorman, K. and Bedrick, S. (2019).
\newblock We need to talk about standard splits.
\newblock In {\em Proceedings of the 57th Annual Meeting of the Association for
  Computational Linguistics (ACL)\/}, pages 2786--2791, Florence, Italy. ACL.

\bibitem[Gururangan {\em et~al.}(2020)Gururangan, Marasovi{\'c}, Swayamdipta,
  Lo, Beltagy, Downey, and Smith]{Gururangan:Marasovi:ACL:2020:Dont}
Gururangan, S.  {\em et~al.} (2020).
\newblock Don{'}t stop pretraining: Adapt language models to domains and tasks.
\newblock In {\em Proceedings of the 58th Annual Meeting of the Association for
  Computational Linguistics (ACL)\/}, pages 8342--8360, Online. ACL.

\bibitem[Haynes {\em et~al.}(2005)Haynes, McKibbon, Wilczynski, Walter, and
  Werre]{Haynes:McKibbon:BMJ:2005:optimal}
Haynes, R.~B.  {\em et~al.} (2005).
\newblock Optimal search strategies for retrieving scientifically strong
  studies of treatment from medline: analytical survey.
\newblock {\em Bmj\/}, {\bf 330}, 1179.

\bibitem[Hofer {\em et~al.}(2018)Hofer, Kormilitzin, Goldberg, and
  Nevado-Holgado]{Hofer:Kormilitzin:2018:FewShot}
Hofer, M.  {\em et~al.} (2018).
\newblock Few-shot learning for named entity recognition in medical text.
\newblock {\em arXiv preprint arXiv:1811.05468\/}.

\bibitem[Khan {\em et~al.}(2020)Khan, Ziyadi, and
  AbdelHady]{Khan:Ziyadi:2020:MtBioNER}
Khan, M.~R.  {\em et~al.} (2020).
\newblock Mt-bioner: Multi-task learning for biomedical named entity
  recognition using deep bidirectional transformers.
\newblock {\em arXiv preprint arXiv:2001.08904\/}.

\bibitem[Kim and Kang(2021)Kim and Kang]{Kim:Kang:2021:BioGeneralize}
Kim, H. and Kang, J. (2021).
\newblock How do your biomedical named entity models generalize to novel
  entities?
\newblock {\em arXiv preprint arXiv:2101.00160\/}.

\bibitem[Lafferty {\em et~al.}(2001)Lafferty, McCallum, and
  Pereira]{Lafferty:MaCallumg:ICML:2001:CRF}
Lafferty, J.~D.  {\em et~al.} (2001).
\newblock Conditional random fields: Probabilistic models for segmenting and
  labeling sequence data.
\newblock In {\em Proceedings of the Eighteenth International Conference on
  Machine Learning\/}, ICML '01, pages 282--289, San Francisco, CA, USA.

\bibitem[Lange {\em et~al.}(2019)Lange, Adel, and
  Str{\"o}tgen]{Lange:Adel:NLNDE:2019:Meddocan}
Lange, L.  {\em et~al.} (2019).
\newblock {NLNDE}: The neither-language-nor-domain-experts' way of spanish
  medical document de-identification.
\newblock In {\em {Proceedings of The Iberian Languages Evaluation Forum
  (IberLEF)}\/}, {CEUR} Workshop Proceedings.

\bibitem[Lange {\em et~al.}(2020)Lange, Dai, Adel, and
  Str{\"{o}}tgen]{Lange:Adel:NLNDE:2020:Cantemist}
Lange, L.  {\em et~al.} (2020).
\newblock {NLNDE} at {CANTEMIST:} neural sequence labeling and parsing
  approaches for clinical concept extraction.
\newblock In {\em {Proceedings of The Iberian Languages Evaluation Forum
  (IberLEF)}\/}, {CEUR} Workshop Proceedings.

\bibitem[Lange {\em et~al.}(2021a)Lange, Adel, and
  Str{\"{o}}tgen]{Lange:Adel:NLNDE:2021:Meddoprof}
Lange, L.  {\em et~al.} (2021a).
\newblock Boosting transformers for job expression extraction and
  classification in a low-resource setting.
\newblock In {\em {Proceedings of The Iberian Languages Evaluation Forum
  (IberLEF)}\/}, {CEUR} Workshop Proceedings.

\bibitem[Lange {\em et~al.}(2021b)Lange, Str{\"o}tgen, Adel, and
  Klakow]{Lange:Stroetgen:EMNLP:2021:Share}
Lange, L.  {\em et~al.} (2021b).
\newblock To share or not to share: {P}redicting sets of sources for model
  transfer learning.
\newblock In {\em Proceedings of the 2021 Conference on Empirical Methods in
  Natural Language Processing (EMNLP)\/}, pages 8744--8753, Online and Punta
  Cana, Dominican Republic. ACL.

\bibitem[Leaman {\em et~al.}(2015)Leaman, Khare, and Lu]{Leaman:Khare:JBI:2015}
Leaman, R.  {\em et~al.} (2015).
\newblock Challenges in clinical natural language processing for automated
  disorder normalization.
\newblock {\em J. Biomed. Inform.}, {\bf 57}, 28--37.

\bibitem[Lee {\em et~al.}(2020)Lee, Yoon, Kim, Kim, Kim, So, and
  Kang]{Lee:Yoon:OxBio:2020:Biobert}
Lee, J.  {\em et~al.} (2020).
\newblock Biobert: a pre-trained biomedical language representation model for
  biomedical text mining.
\newblock {\em Bioinformatics\/}, {\bf 36}, 1234—1240.

\bibitem[Lee {\em et~al.}(2018)Lee, Dernoncourt, and
  Szolovits]{Lee:Dernoncourt:LREC:2018:Transfer}
Lee, J.~Y.  {\em et~al.} (2018).
\newblock Transfer learning for named-entity recognition with neural networks.
\newblock In {\em Proceedings of the Eleventh International Conference on
  Language Resources and Evaluation (LREC)\/}, Miyazaki, Japan. European
  Language Resources Association.

\bibitem[Lima-López {\em et~al.}(2021)Lima-López, ans Antonio
  Miranda-Escalada, Brivá-Iglesias, and
  Krallinger]{Lima:Farre:MEDDOPROF:2021:Overview}
Lima-López, S.  {\em et~al.} (2021).
\newblock Nlp applied to occupational health: Meddoprof shared task at iberlef
  2021 on automatic recognition, classification and normalization of
  professions and occupations from medical texts.
\newblock In {\em Proceedings of the Iberian Languages Evaluation Forum
  (IberLEF)\/}, {CEUR} Workshop Proceedings.

\bibitem[Luoma and Pyysalo(2020)Luoma and
  Pyysalo]{Luoma:Ppysalo:COLING:2020:ContextNER}
Luoma, J. and Pyysalo, S. (2020).
\newblock Exploring cross-sentence contexts for named entity recognition with
  {BERT}.
\newblock In {\em Proceedings of the 28th International Conference on
  Computational Linguistics (COLING)\/}, pages 904--914, Barcelona, Spain
  (Online). International Committee on Computational Linguistics.

\bibitem[Marimon {\em et~al.}(2019)Marimon, Gonzalez-Agirre, Intxaurrondo,
  Rodríguez, Lopez~Martin, Villegas, and
  Krallinger]{Marimon:Gonzalez:MEDDOCAN:2019:Overview}
Marimon, M.  {\em et~al.} (2019).
\newblock Automatic de-identification of medical texts in spanish: the meddocan
  track, corpus, guidelines, methods and evaluation of results.
\newblock In {\em Proceedings of the Iberian Languages Evaluation Forum
  (IberLEF)\/}. CEUR Workshop Proceedings.

\bibitem[Miranda-Escalada {\em et~al.}(2020)Miranda-Escalada, Farré, and
  Krallinger]{Miranda:Farre:CANTEMIST:2020:Overview}
Miranda-Escalada, A.  {\em et~al.} (2020).
\newblock Named entity recognition, concept normalization and clinical coding:
  Overview of the cantemist track for cancer text mining in spanish, corpus,
  guidelines, methods and results.
\newblock In {\em {Proceedings of the Iberian Languages Evaluation Forum
  (IberLEF)}\/}, {CEUR} Workshop Proceedings.

\bibitem[Mulyar {\em et~al.}(2021)Mulyar, Uzuner, and
  McInnes]{Mulyar:Uzuner:2021:JAIMA:MtClinicalBert}
Mulyar, A.  {\em et~al.} (2021).
\newblock Mt-clinical bert: scaling clinical information extraction with
  multitask learning.
\newblock {\em J. Am. Med. Inform. Assoc.}, {\bf 28}, 2108--2115.

\bibitem[Naseem {\em et~al.}(2021)Naseem, Khushi, Reddy, Rajendran, Razzak, and
  Kim]{Aaseem:Kushi:IJCNN:2021:Bioalbert}
Naseem, U.  {\em et~al.} (2021).
\newblock Bioalbert: A simple and effective pre-trained language model for
  biomedical named entity recognition.
\newblock In {\em 2021 International Joint Conference on Neural Networks
  (IJCNN)\/}, pages 1--7. IEEE.

\bibitem[Peng {\em et~al.}(2019)Peng, Yan, and
  Lu]{Peng:Yan:BioNLP:2019:TransferBio}
Peng, Y.  {\em et~al.} (2019).
\newblock Transfer learning in biomedical natural language processing: An
  evaluation of {BERT} and {ELM}o on ten benchmarking datasets.
\newblock In {\em Proceedings of the 18th BioNLP Workshop and Shared Task
  (BioNLP)\/}, pages 58--65, Florence, Italy. ACL.

\bibitem[Peters {\em et~al.}(2018)Peters, Neumann, Iyyer, Gardner, Clark, Lee,
  and Zettlemoyer]{Peters:Neumann:NAACL:2018:Deep}
Peters, M.~E.  {\em et~al.} (2018).
\newblock Deep contextualized word representations.
\newblock In {\em Proceedings of the 2018 Conference of the North {A}merican
  Chapter of the Association for Computational Linguistics: Human Language
  Technologies (NAACL-HLT)\/}, pages 2227--2237, New Orleans, Louisiana. ACL.

\bibitem[Phan {\em et~al.}(2019)Phan, Sun, and
  Tay]{Phan:Sun:ACL:2019:ContextRepr}
Phan, M.~C.  {\em et~al.} (2019).
\newblock Robust representation learning of biomedical names.
\newblock In {\em Proceedings of the 57th Annual Meeting of the Association for
  Computational Linguistics (ACL)\/}, pages 3275--3285, Florence, Italy. ACL.

\bibitem[Schweter and Akbik(2020)Schweter and Akbik]{Schweter:Akbik:2020:flert}
Schweter, S. and Akbik, A. (2020).
\newblock Flert: Document-level features for named entity recognition.
\newblock {\em arXiv preprint arXiv:2011.06993\/}.

\bibitem[Stubbs {\em et~al.}(2015)Stubbs, Kotfila, and
  Uzuner]{Stubbs:Kotfile:i2b2:2014:Overview}
Stubbs, A.  {\em et~al.} (2015).
\newblock Automated systems for the de-identification of longitudinal clinical
  narratives: Overview of 2014 i2b2/uthealth shared task track 1.
\newblock {\em J. Biomed. Inform.}, {\bf 58}, 11--19.

\bibitem[Sun {\em et~al.}(2013)Sun, Rumshisky, and
  Uzuner]{Sun:Rumshisky:i2b2:2012:Overview}
Sun, W.  {\em et~al.} (2013).
\newblock Evaluating temporal relations in clinical text: 2012 i2b2 challenge.
\newblock {\em J. Am. Med. Inform. Assoc.}, {\bf 20}, 806--813.

\bibitem[Uzuner {\em et~al.}(2007)Uzuner, Luo, and
  Szolovits]{Uzumer:Luo:i2b2:2006:Overview}
Uzuner, {\"O}.  {\em et~al.} (2007).
\newblock Evaluating the state-of-the-art in automatic de-identification.
\newblock {\em J. Am. Med. Inform. Assoc.}, {\bf 14}, 550--563.

\bibitem[Uzuner {\em et~al.}(2011)Uzuner, South, Shen, and
  DuVall]{Uzuner:South:i2b2:2010:Overview}
Uzuner, {\"O}.  {\em et~al.} (2011).
\newblock 2010 i2b2/va challenge on concepts, assertions, and relations in
  clinical text.
\newblock {\em J. Am. Med. Inform. Assoc.}, {\bf 18}, 552--556.

\bibitem[Villegas {\em et~al.}(2018)Villegas, Intxaurrondo, Gonzalez-Agirre,
  Marimon, and Krallinger]{Villegas:Intxaurrondo:BIO:2018:Mespen}
Villegas, M.  {\em et~al.} (2018).
\newblock The mespen resource for english-spanish medical machine translation
  and terminologies: census of parallel corpora, glossaries and term
  translations.
\newblock {\em LREC MultilingualBIO\/}.

\bibitem[Wang {\em et~al.}(2018)Wang, Wang, Rastegar-Mojarad, Moon, Shen,
  Afzal, Liu, Zeng, Mehrabi, and Sohn]{Wang:Wang:JBI:2018:Review}
Wang, Y.  {\em et~al.} (2018).
\newblock Clinical information extraction applications: a literature review.
\newblock {\em J. Biomed. Inform.}, {\bf 77}, 34--49.

\bibitem[Weber {\em et~al.}(2019)Weber, Münchmeyer, Rocktäschel, Habibi, and
  Leser]{Weber:Muenchmeyer:OxBio:2019:HUNER}
Weber, L.  {\em et~al.} (2019).
\newblock {HUNER: improving biomedical NER with pretraining}.
\newblock {\em Bioinformatics\/}, {\bf 36}, 295--302.

\bibitem[Weber {\em et~al.}(2021)Weber, Sänger, Münchmeyer, Habibi, Leser,
  and Akbik]{Weber:Saenger:OxBio:2021:HunFlair}
Weber, L.  {\em et~al.} (2021).
\newblock {HunFlair: an easy-to-use tool for state-of-the-art biomedical named
  entity recognition}.
\newblock {\em Bioinformatics\/}, {\bf 37}, 2792--2794.

\end{thebibliography}

\end{document}